\documentclass[11pt,a4paper]{article}
\usepackage[hyperref]{emnlp2020}
\usepackage{times}
\usepackage{latexsym}

\usepackage{microtype}

\usepackage{enumitem}
\newlist{compactitem}{itemize}{3} 
\setlist[compactitem]{label=\textbullet, nosep}

\usepackage{amssymb}
\usepackage{pifont}
\newcommand{\cmark}{\ding{51}}%
\newcommand{\xmark}{\ding{55}}%

\usepackage{pbox}
\usepackage{hyperref}
\usepackage{url}
\usepackage{comment}
\usepackage{subfig}
\usepackage[flushleft]{threeparttable}
\usepackage{graphicx}
\usepackage{amsmath}
\usepackage{subfig}
\usepackage{tabularx}
\usepackage{multirow, makecell}
\usepackage{algorithm}
\usepackage{xcolor}
\usepackage{booktabs}
\usepackage{paralist}

\aclfinalcopy 

\definecolor{MyColor}{RGB}{50, 100, 250}
\definecolor{Orange}{RGB}{244, 101, 66}
\definecolor{Red}{RGB}{255, 0, 0}
\definecolor{Green}{RGB}{0, 255, 0}
\definecolor{Blue}{RGB}{0, 0, 255}

\newcommand{\mytilde}{\raise.17ex\hbox{$\scriptstyle\mathtt{\sim}$}}

\title{PolicyQA: A Reading Comprehension Dataset for Privacy Policies}

\author{
Wasi Uddin Ahmad\thanks{~~Equal contribution.} \\
University of California, Los Angeles \\
\texttt{wasiahmad@ucla.edu} \\
\And
Jianfeng Chi\footnotemark[1] \\
University of Virginia \\
\texttt{jc6ub@virginia.edu} \\
\AND 
Yuan Tian \\
University of Virginia \\
\texttt{yuant@virginia.edu} \\
\And
Kai-Wei Chang \\
University of California, Los Angeles \\
\texttt{kwchang.cs@ucla.edu} \\
}

\date{}

\begin{document}
\maketitle

\begin{abstract}
Privacy policy documents are long and verbose.
A question answering (QA) system can assist users in finding the information that is relevant and important to them.
Prior studies in this domain frame the QA task as retrieving the most relevant text segment or a list of sentences from the policy document given a question.
On the contrary, we argue that providing users with a short text span from policy documents reduces the burden of searching the target information from a lengthy text segment.
In this paper, we present PolicyQA, a dataset that contains 25,017 reading comprehension style examples curated from an existing corpus of 115 website privacy policies.
PolicyQA provides 714 human-annotated questions written for a wide range of privacy practices.
We evaluate two existing neural QA models and perform rigorous analysis to reveal the advantages and challenges offered by PolicyQA.
\end{abstract}

\vspace{0.5mm}
\section{Introduction}
Security and privacy policy documents describe how an entity collects, maintains, uses, and shares users' information.
Users need to read the privacy policies of the websites they visit or the mobile applications they use and know about their privacy practices that are pertinent to them.
However, prior works suggested that people do not read privacy policies because they are long and complicated \cite{mcdonald2008cost}, and confusing \cite{reidenberg2016ambiguity}.
Hence, giving users access to a question answering system to search for answers from long and verbose policy documents can help them better understand their rights.

In recent years, we have witnessed noteworthy progress in developing question answering (QA) systems with a colossal effort to benchmark high-quality, large-scale datasets for a few application domains (e.g., Wikipedia, news articles). 
However, annotating large-scale QA datasets for domains such as security and privacy is challenging as it requires expert annotators (e.g., law students).
Due to the difficulty of annotating policy documents at scale, the only available QA dataset is PrivacyQA \cite{Ravichander2019Question} on privacy policies for 35 mobile applications.

\begin{table}[t]
\centering
\vspace{4mm}
\resizebox{\linewidth}{!}{%
\def\arraystretch{1.5}%
\begin{tabular}{p{0.99\linewidth}}
\hline
Website: Amazon.com \\
\hline

Information You Give Us: We receive and store any {\color{red} information you enter on our Web site} or give us in any other way. Click here to see ...

\medskip
\underline{Question}. How do you collect my information? \\
\vspace{-6mm}
{\color{red} information you enter on our Web site} \\

\hline

Promotional Offers: Sometimes we send offers to selected groups of Amazon.com customers on behalf of other businesses. When we do this, {\color{blue}we do not give that business your name and address}. If you do not want to receive such offers, ...

\medskip
\underline{Question}. Is my information shared with others? \\
\vspace{-6mm}
{\color{blue}we do not give that business your name and address} \\

\hline

\end{tabular}
}
\vspace{-2mm}
\caption{Question-answer pairs that we collect from OPP-115 \cite{wilson2016creation} dataset. The evidence spans are highlighted in color and they are used to form the question-answer pairs.}
\label{table:example}
\end{table}
\begin{table*}[!ht]
\centering
\begin{tabular}{l|l|l}
\hline
& PolicyQA (This work) & PrivacyQA \\
\hline
\hline
Source & Website privacy policies & Mobile application privacy policies  \\
\# Policies & 115 & 35 \\
\# Questions & 714 & 1,750 \\
\# Annotations & 25,017 & 3,500 \\
Question annotator & Domain experts & Mechanical Turkers \\
Form of QA & Reading comprehension & Sentence selection \\
Answer type & A sequence of words & A list of sentences \\
\hline
\end{tabular}
\vspace{-2mm}
\caption{Comparison of PolicyQA and PrivacyQA.}
\label{table:privacyqa_vs_policyqa}
\end{table*}


An essential characteristic of policy documents is that they are well structured as they are written by following guidelines set by the policymakers.
Besides, due to the homogeneous nature of different entities (e.g., Amazon, eBay), their privacy policies have a similar structure.
Therefore, we can exploit the document structure (meta data) to form examples from existing corpora.
In this paper, we present PolicyQA, a reading comprehension style question answering dataset with 25,017 question-passage-answer triples associated with text segments from privacy policy documents.
PolicyQA consists of 714 questions on 115 website privacy policies and is curated from an existing corpus, OPP-115 \cite{wilson2016creation}.
Table \ref{table:example} presents a couple of examples from PolicyQA.

In contrast to PrivacyQA \cite{Ravichander2019Question} that focuses on extracting long text spans from policy documents, we argue that highlighting a shorter text span in the document facilitates the users to zoom into the policy and identify the target information quickly.
To enable QA models to provide such short answers, PolicyQA provides examples with an average answer length of 13.5 words (in comparison, the PrivacyQA benchmark has examples with an average answer length of 139.6 words). 
We present a comparison between PrivacyQA and PolicyQA in Table \ref{table:privacyqa_vs_policyqa}.

In this work, we present two strong neural baseline models trained on PolicyQA and perform a thorough analysis to shed light on the advantages and challenges offered by the proposed dataset.
The data and the implemented baseline models are made publicly available.\footnote{https://github.com/wasiahmad/PolicyQA}

\section{Dataset}

PolicyQA consists question-passage-answer triples, curated from OPP-115  \cite{wilson2016creation}.
OPP-115 is a corpus of 115 website privacy policies (3,792 segments), manually annotated by skilled annotators following the annotation schemes predefined by domain experts.
The annotation schemes are composed of 10 data practice categories (e.g., \textit{First Party Collection/Use}, \textit{Third Party Sharing/Collection}, \textit{User Choice/Control} etc.).
The data practices are further categorized into a set of practice attributes (e.g., \textit{Personal Information Type}, \textit{Purpose}, \textit{User Type} etc.).
Each practice attribute is associated with a predefined set of values.
In the Appendix (in Table \ref{table:attr_val_pairs}), we list all the attributes under the \textit{First Party Collection/Use} category.


\begin{table}[t]
\vspace{2mm}
\centering
\resizebox{\linewidth}{!}{%
\def\arraystretch{1.1}%
\begin{tabular}{p{0.99\linewidth}}
\hline
``Practice'': First Party Collection/Use \\
``Attribute'': Purpose \\
``value'': ``Additional service/feature'' \\
``startIndexInSegment'': 360 \\ 
``endIndexInSegment'': 387 \\
``selectedText'': ``responding to your requests''
\\ \hline
``Practice'': Third Party Sharing/Collection \\
``Attribute'': Third Party Entity \\
``value'': ``Unnamed third party'' \\
``startIndexInSegment'': 573 \\
``endIndexInSegment'': 596 \\
``selectedText'': ``Third-Party Advertisers''
\\ 
\hline
\end{tabular}
}
\vspace{-2mm}
\caption{Sample span annotations from OPP-115 associated with a segment of \emph{Amazon.com} privacy policy.
}
\label{table:practice_ex}
\end{table}

In total, OPP-115 contains 23,000 data practices, 128,000 practice attributes, and 103,000 annotated text spans. 
Each text span belongs to a policy segment, and OPP-115 provides its character-level start and end indices.
We provide an example in Table \ref{table:practice_ex}.
We use the annotated spans, corresponding policy segments, and the associated \emph{\{Practice, Attribute, Value\}} triples to form PolicyQA examples.
We exclude the spans with practices labeled as ``Other'' and the values labeled as ``Unspecified''.
Next, we describe the question annotation process.

\begin{figure*}
\captionsetup[subfigure]{labelformat=empty}
\centering
\hspace{-5mm}
\subfloat[(a) PolicyQA (This work) \label{fig2:sub1}]
{
\includegraphics[width=.50\linewidth]{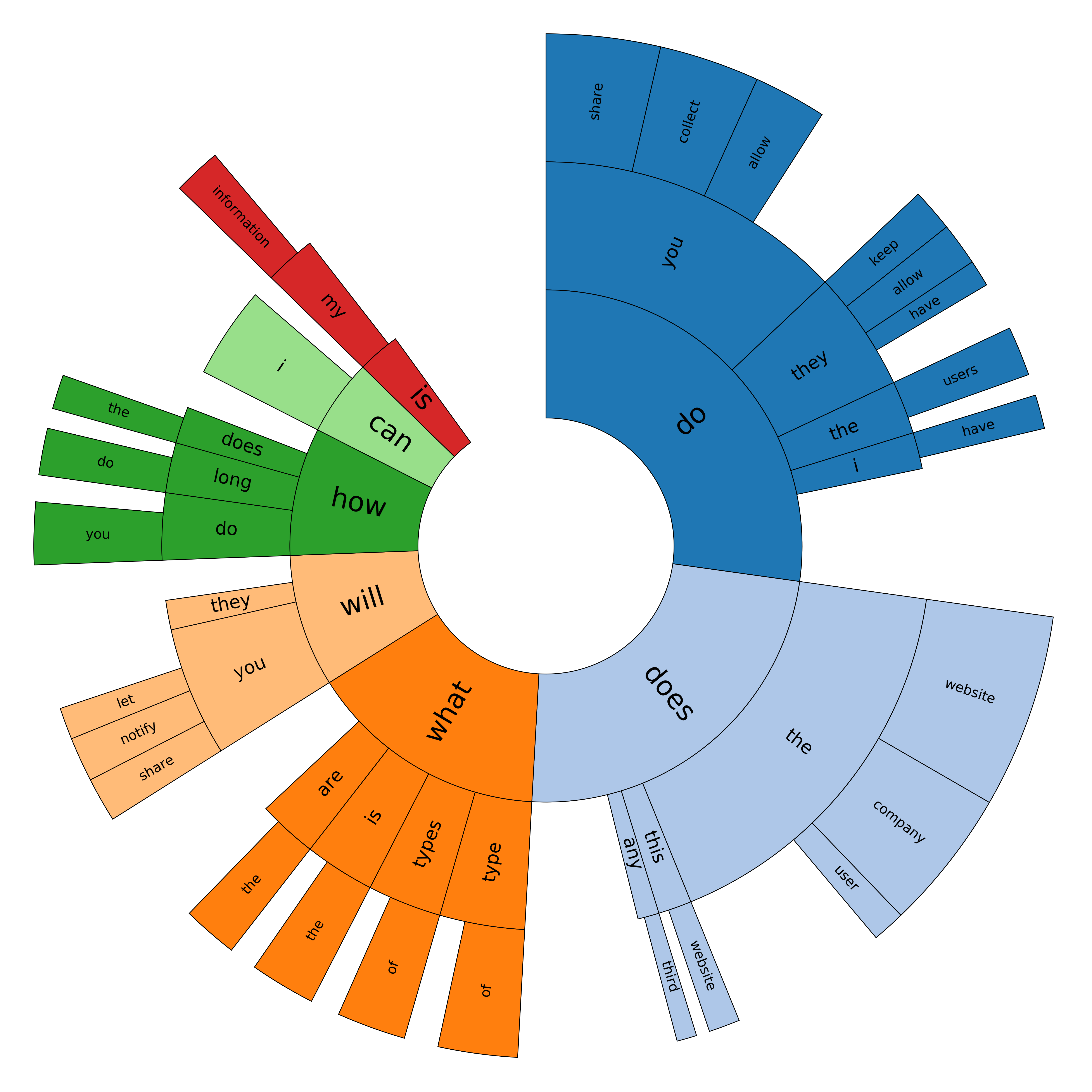}
}
\hfil
\subfloat[(b) PrivacyQA \label{fig2:sub2}]
{
\includegraphics[width=.50\linewidth]{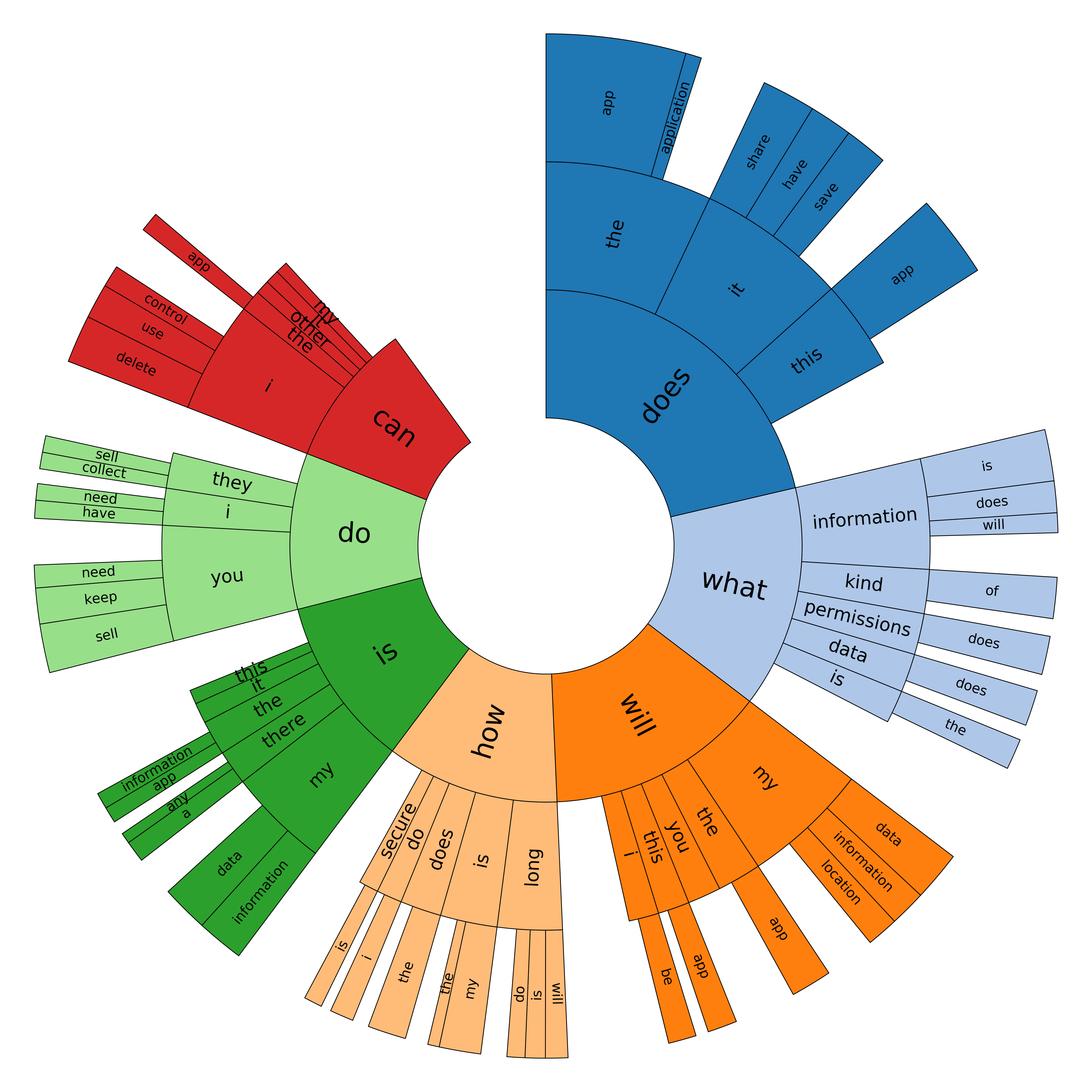}
}
\caption{Distribution of trigram prefixes of questions in (a) PolicyQA and (b) PrivacyQA.}
\vspace{2mm}
\label{figure:query_position}
\end{figure*}
\begin{table*}[t]
\centering
\resizebox{\linewidth}{!}{%
\begin{tabular}{l|l|l}
\hline
Privacy Practice & Proportion & Example Question From PolicyQA \\ 
\hline\hline
First Party Collection/Use &  44.4 \% & Why do you collect my data?
\\
Third Party Sharing/Collection & 34.1 \% & Do they share my information with others?  \\
Data Security & 2.2 \% & Do you use encryption to secure my data?  \\
Data Retention & 1.7 \% & How long they will keep my data?  \\
User Access, Edit and Deletion & 3.1 \% & Will you let me access and edit my data?  \\
User Choice/Control & 11.0 \% & What use of information does the user choice apply to?  \\
Policy Change & 1.9 \% & How does the website notify about policy changes?  \\
International and Specific Audiences & 1.5 \% & What is the company's policy towards children?  \\
Do Not Track & 0.1 \% & Do they honor the user's do not track preference?  \\
\hline
\end{tabular}
}
\vspace{-2mm}
\caption{
OPP-115 categories of the questions in the PolicyQA dataset.
}
\label{table:ques_dist}
\end{table*}

\smallskip\noindent
\textbf{Question annotations.}
Two skilled annotators manually annotate the questions.
During annotation, the annotators are provided with the triple \{Practice, Attribute, Value\}, and the associated text span.
For example, given the triple \{\textit{First Party Collection/Use}, \textit{Personal Information Type}, \textit{Contact}\} and the associated text span ``\emph{name, address, telephone number, email address}'', the annotators created questions, such as, (1) \textit{What type of contact information does the company collect?}, (2) \textit{Will you use my contact information?}, etc.

For a specific triple, the process is repeated for 5-10 randomly chosen samples to form a list of questions.
We randomly assign a question from this list to the examples associated with the triple that were not chosen during the sampling process.
In total, we considered 258 unique triples and created 714 individual questions.
In Table \ref{table:ques_dist}, we provide an example question for each practice category.
Also, we compare the distribution of questions' trigram prefixes in PolicyQA (Figure \ref{fig2:sub1}) with PrivacyQA (Figure \ref{fig2:sub2}). 
It is important to note that, PolicyQA questions are written in a generic fashion to become applicable for text spans, associated with the same practice categories.
Therefore, PolicyQA questions are less diverse than PrivacyQA questions.

We split OPP-115 into 75/20/20 policies to form training, validation, and test examples, respectively.
Table~\ref{table:statistics} summarizes the data statistics.

\section{Experiment}
In this section, we evaluate two neural question answering (QA) models on PolicyQA and present the findings from our analysis.

\smallskip\noindent
\textbf{Baselines.}
PolicyQA frames the QA task as predicting the answer span that exists in the given policy segment.
Hence, we consider two existing neural approaches from literature as baselines for PolicyQA.
The first model is \textbf{BiDAF} \cite{seo2016bidirectional} that uses a bi-directional attention flow mechanism 
to extract the evidence spans.
The second baseline is based on \textbf{BERT} \cite{devlin2019bert}
with two linear classifiers to predict the boundary of the evidence, as suggested in the original work.

\begin{table}[t]
\centering
\resizebox{\linewidth}{!}{%
\begin{tabular}{l|l|l|l}
\hline
Dataset & Train & Valid & Test \\ 
\hline\hline
\# Examples &  17,056 & 3,809 & 4,152  \\
\# Policies & 75 & 20 & 20  \\
\# Questions & 693 & 568 & 600  \\
\# Passages & 2,137 & 574 & 497 \\
\hline
Avg. question length & 11.2 & 11.2 & 11.2  \\
Avg. passage length & 106.0 & 96.6 & 119.1  \\
Avg. answer length & 13.3 & 12.8 & 14.1 \\
\hline
\end{tabular}
}
\vspace{-2mm}
\caption{
Statistics of the PolicyQA dataset.
}
\label{table:statistics}
\end{table}

\begin{table}[t]
\centering
\resizebox{\linewidth}{!}{%
\small
\def\arraystretch{1.1}%
\begin{tabular}{@{\hskip 0.05in}c@{\hskip 0.05in}|@{\hskip 0.05in}c@{\hskip 0.05in}|c@{\hskip 0.1in} c|c@{\hskip 0.1in} c}
\hline
\multirowcell{2}{Fine- \\ tuning} & \multirowcell{2}{SQuAD \\ Pre-training}  & \multicolumn{2}{c|}{Valid} & \multicolumn{2}{c}{Test}  \\ 
\cline{3-6}
& & EM & F1 & EM & F1 \\ 
\hline\hline
\multicolumn{6}{l}{BiDAF} \\
\hline
\xmark & \xmark & 25.1 & 52.3 & 22.0 & 48.0 \\
\xmark & \cmark & 26.7 & 53.7 & 23.3 & 49.5 \\
\cmark & \xmark & 27.9 & 57.2 & 24.4 & 52.8 \\
\hline
\multicolumn{6}{l}{BERT-base} \\
\hline
\xmark & \xmark & 30.5 & 59.4 & 28.1 & 55.6 \\
\xmark & \cmark & 30.5 & 60.2 & 28.0 & 56.2 \\
\cmark & \xmark & {\bf 32.8} & 60.9 & 28.6 & {\bf 56.6} \\
\cmark & \cmark & 32.7 & {\bf 61.2} & {\bf 29.5} & {\bf 56.6} \\
\hline
\end{tabular}
}
\vspace{-2mm}
\caption{Performance of baselines on PolicyQA.
The bold face values indicate the best performances.
}
\label{table:ext_results}
\end{table}

\smallskip\noindent
\textbf{Implementation.}
PolicyQA has a similar setting as SQuAD \cite{rajpurkar2016squad}. Therefore, we pre-train the QA models using their default settings on the SQuAD dataset.
Besides, we consider leveraging unlabeled privacy policies in fine-tuning the models, as noted below.

\smallskip\noindent
$\bullet$ \textbf{Fine-tuning.}
We train word embeddings using \emph{fastText} \cite{bojanowski2017enriching} based on a corpus of 130,000 privacy policies (137M words) collected from apps on the Google Play Store.\footnote{We thank the authors of \cite{harkous2018polisis} for sharing the 130,000 privacy policies.}
These word embeddings are used as fixed word representations in BiDAF while training on PolicyQA.
Similarly, to adapt BERT to the privacy domain, we first fine-tune BERT using masked language modeling \cite{devlin2019bert} based on the privacy policies and then train on PolicyQA.

\smallskip\noindent
$\bullet$ \textbf{No fine-tuning.}
In this setting, we use the publicly available \emph{fastText} \cite{bojanowski2017enriching} embeddings with BiDAF, and the BERT model is not fine-tuned on those privacy policies.

We adopt the default model architecture and optimization setup for the baseline methods.
We detail the hyper-parameters in Appendix (in Table \ref{table:hyperparameters}).


\smallskip\noindent
\textbf{Evaluation.}
Following \citet{rajpurkar2016squad}, we use \emph{exact match} (EM) and \emph{F1 score} to evaluate the model's accuracy.

\begin{table}[!ht]
\centering
\begin{tabular}{l|c|c|c|c}
\hline
\multirow{2}{*}{BERT Size} & \multicolumn{2}{c|}{Valid} & \multicolumn{2}{c}{Test}  \\ 
\cline{2-5}
& EM & F1 & EM & F1 \\ 
\hline\hline
Tiny & 21.0 & 47.1 & 15.5 & 39.9 \\
Mini & 26.5 & 55.2 & 22.8 & 49.8 \\
Small & 28.4 & 57.2 & 24.6 & 52.3 \\
Medium & {\bf 31.1} & 59.1 & 25.2 & 53.5 \\
Base & 30.5 & {\bf 59.4} & {\bf 28.1} & {\bf 55.6} \\
\hline
\end{tabular}
\vspace{-2mm}
\caption{Performance of different sized QA models. 
}
\label{table:bert_results}
\end{table}

\begin{table}[!ht]
\centering
\resizebox{\linewidth}{!}{
\begin{tabular}{@{\hskip 0.05in}l@{\hskip 0.05in}|c|c|c} 
\hline
& $|\text{ans}|$ & EM & F1 \\ 
\hline\hline
Third Party Sharing/Collection & 9.3 & 35.0 & 60.2 \\ 
First Party Collection/Use & 10.1 & 28.3 & 55.7 \\
Data Retention & 10.6 & 29.1 & 55.9 \\ 
User Choice/Control & 11.0 & 24.3 & 53.2 \\ 
User Access, Edit and Deletion & 12.2 & 21.6 & 51.5 \\ 
Policy Change & 14.6 & 43.4 & 67.7 \\ 
Do Not Track & 30.9 & 37.5 & 69.2 \\
Data Security & 34.6 & 24.4 & 54.3 \\ 
Intl. and Specific Audiences & 52.8 & 5.3 & 43.1 \\ 
\hline
\end{tabular}
}
\vspace{-2mm}
\caption{Test performance breakdown of BERT-base model for privacy practice categories, sorted by the average answer length as indicated by $|\text{ans}|$.
}
\label{table:breakdown_result}
\end{table}

\subsection{Results and Analysis}

The experimental results are presented in Table \ref{table:ext_results}.
Overall, the BERT-base methods outperform the BiDAF models by 6.1\% and 7.6\% in terms of EM and F1 score (on the test split), respectively.


\smallskip\noindent
\textbf{Impact of fine-tuning.}
Table \ref{table:ext_results} demonstrates that the fine-tuning step improves the downstream task performance.
For example, BERT-base performance is improved by 0.5\% and 1.0\% EM and F1 score, respectively, on the test split.
This result encourages to train/fine-tune BERT on a larger collection of security and privacy documents.

\smallskip\noindent
\textbf{Impact of SQuAD pre-training.}
Given a small number of training examples, it is challenging to train deep neural models. 
Hence, we pre-train the extractive QA models on SQuAD \cite{rajpurkar2016squad} and then fine-tune on PolicyQA.
The additional pre-training step improves performance.
For example, in \emph{no fine-tuning} setting, BiDAF, and BERT-base improve the performance by 1.5\% and 0.6\% F1 score, respectively (on the test split).

\begin{figure*}[t]
\captionsetup[subfigure]{labelformat=empty}
\centering
\hspace{-4mm}
\subfloat[(a) \label{fig:qtype_anal}]
{
\includegraphics[width=0.49\textwidth]{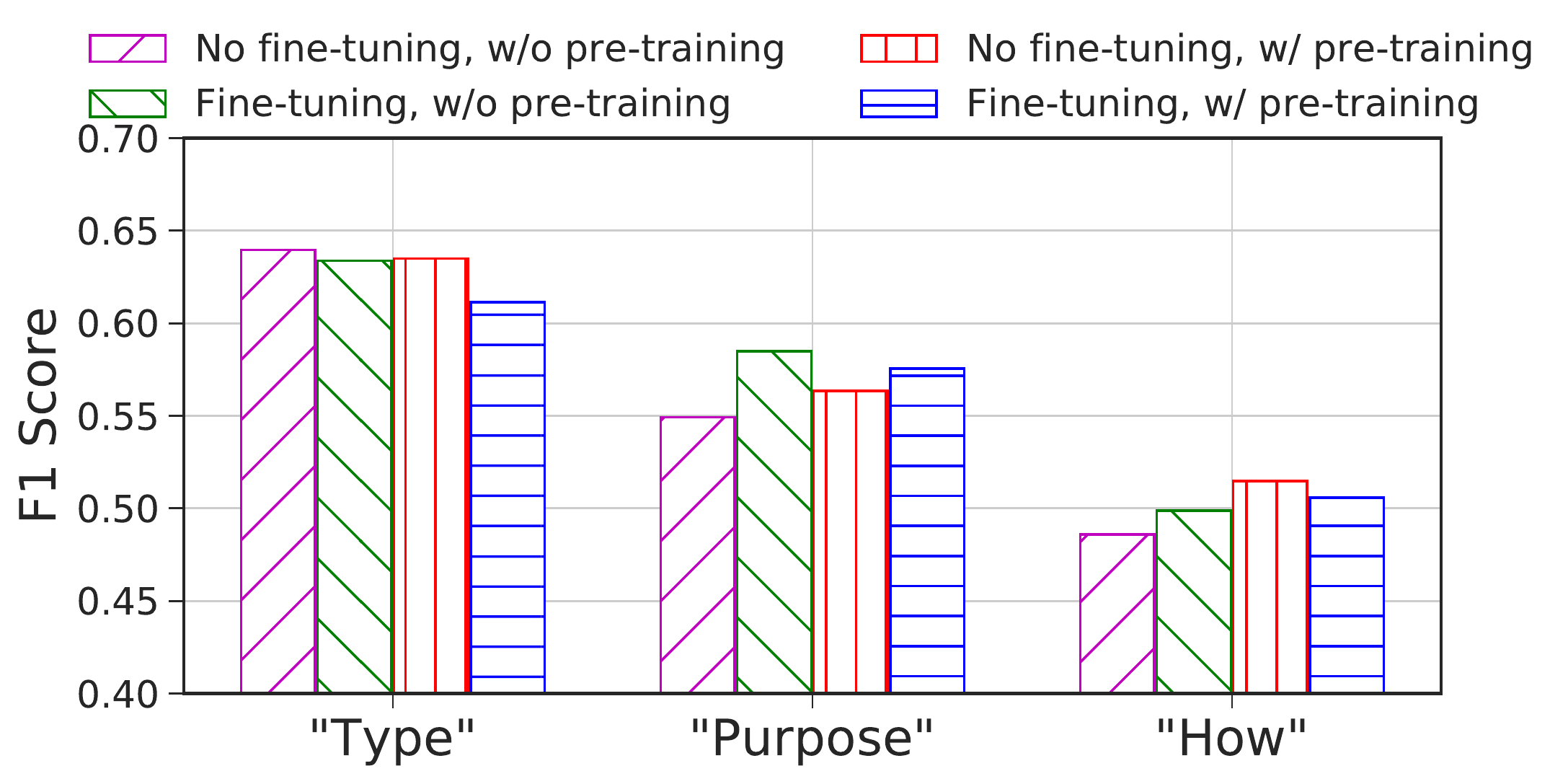}
}
\hfil
\subfloat[(b) \label{fig:atype_anal}]
{
\includegraphics[width=0.49\textwidth]{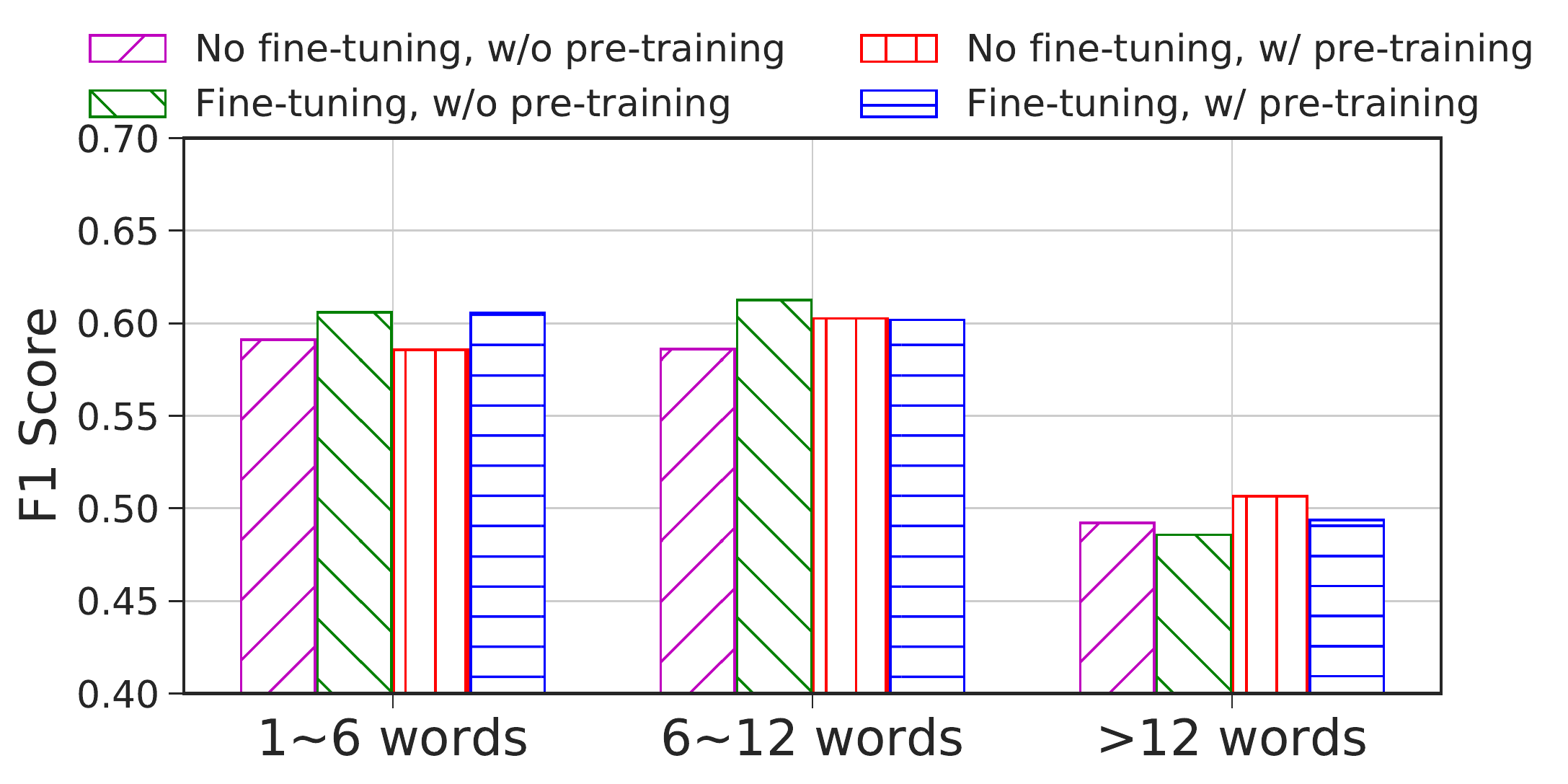}
}
\vspace{-2mm}
\caption{BERT-base model's performance on (a) the three most frequent attributes of ``First Party Collection/Use'' and ``Third Party Sharing/Collection'' practice categories, and (b) questions with different answer lengths.}
\label{figure:aba_anlaysis}
\end{figure*}

\smallskip\noindent
\textbf{Impact of model size.}
We experiment with different sized BERT models \cite{turc2019well} and the results in Table \ref{table:bert_results} shows that the performance improves with increased model size. 
The results also indicate that PolicyQA is a challenging dataset, and hence, a larger model performs better.


\smallskip\noindent
\textbf{Analysis.}
We breakdown the test performance of the BERT-base method to examine the model performance across practice categories.
The results are presented in Table \ref{table:breakdown_result}.
We see the model performs comparably on the three most frequent categories (comprise 89.5\% of the total examples).

We further analyze the performance on questions associated with (1) the top three frequent attributes for the two most frequent practice categories, and (2) different answer lengths.
The results are presented in Figure \ref{fig:qtype_anal} and \ref{fig:atype_anal}. 
Our findings are (1) shorter evidence spans (e.g., evidence spans for \emph{Personal Information Type} questions) are easier to extract than longer spans; 
and (2) SQuAD pre-training helps more in extracting shorter evidence spans.
Leveraging diverse extractive QA resources may reduce the length bias and boost the QA performance on privacy policies.

\section{Related Work}
The \emph{Usable Privacy Project} \cite{sadeh2013usable} has made several attempts to automate the analysis of privacy policies \cite{wilson2016creation, zimmeck2019maps}. 
Noteworthy works include identification of policy segments commenting on specific data practices \cite{wilson2016crowdsourcing}, extraction of opt-out choices, and their provisions in policy text \cite{sathyendra2016automatic, sathyendra2017identifying}, and others \cite{bhatia2015towards, bhatia2016automated}.
\citet{kaur2018comprehensive} used a keyword-based technique to compare online privacy policies. 
Natural language processing (NLP) techniques such as text alignment~\cite{liu2014step, ramanath2014unsupervised}, text classification~\cite{harkous2018polisis, zimmeck2019maps, wilson2016creation} and question answering \cite{shvartzshanider2018recipe, harkous2018polisis, Ravichander2019Question} has been studied in prior works to facilitate privacy policy analysis. 

Among the question answering (QA) methods, \citet{harkous2018polisis} framed the task as retrieving the most relevant policy segments as an answer, while \citet{Ravichander2019Question} presented a dataset and models to answer questions with a list of sentences.
In comparison to the prior QA approaches, we encourage developing QA systems capable of providing precise answers by using PolicyQA.

\section{Conclusion}
This work proposes PolicyQA, a reading comprehension style question answering (QA) dataset.
PolicyQA can contribute to the development of QA systems in the security and privacy domain that have a sizeable real-word impact. 
We evaluate two strong neural baseline methods on PolicyQA and provide thorough ablation analysis to reveal important considerations that affect answer span prediction.
In our future work, we want to explore how transfer learning can benefit question answering in the security and privacy domain.

\section*{Acknowledgments}
This work was supported in part by National Science Foundation Grant OAC 1920462.

\bibliography{bib_short}
\bibliographystyle{acl_natbib}

\cleardoublepage










\appendix

\addcontentsline{toc}{section}{Appendices}
\renewcommand{\thesubsection}{\Alph{subsection}}



\begin{table*}[!ht]
\centering
\resizebox{\linewidth}{!}{%
\small
\def\arraystretch{1.5}%
\begin{tabular}{l | l}
\hline
Attribute & Values \\ 
\hline\hline
Does/Does Not & \multicolumn{1}{m{10cm}}{Does; Does Not} 
\\ \hline
Collection Mode & \multicolumn{1}{m{10cm}}{Explicit; Implicit; Unspecified} 
\\ \hline
Action First-Party & \multicolumn{1}{m{10cm}}{Collect on website; Collect in mobile app; Collect on mobile website; Track user on other websites; Collect from user on other websites; Receive from other parts of company/affiliates; Receive from other service/third-party (unnamed); Receive from other service/third-party (named); Other; Unspecified} 
\\ \hline
Identifiability & \multicolumn{1}{m{10cm}}{Identifiable; Aggregated or anonymized; Other; Unspecified} 
\\ \hline
Personal Information Type & \multicolumn{1}{m{10cm}}{Financial; Health; Contact; Location; Demographic; Personal identifier; User online activities; User profile; Social media data; IP address and device IDs; Cookies and tracking elements; Computer information; Survey data; Generic personal information; Other; Unspecified} 
\\ \hline
Purpose & \multicolumn{1}{m{10cm}}{Basic service/feature; Additional service/feature; Advertising; Marketing; Analytics/Research; Personalization/Customization; Service Operation and Security; Legal requirement; Merger/Acquisition; Other; Unspecified} 
\\
\hline
User Type & \multicolumn{1}{m{10cm}}{User without account; User with account; Other; Unspecified} 
\\ \hline
Choice Type & \multicolumn{1}{m{10cm}}{Dont use service/feature; Opt-in; Opt-out link; Opt-out via contacting company; First-party privacy controls; Third-party privacy controls; Browser/device privacy controls; Other; Unspecified} 
\\ \hline
Choice Scope & \multicolumn{1}{m{10cm}}{Collection; Use; Both; Unspecified} 
\\ \hline
\end{tabular}
}
\caption{The attributes and their values for the \emph{First Party Collection/Use} data practice category. We do not consider the data practices associated with ``Unspecified'' values.}
\label{table:attr_val_pairs}
\end{table*}

\begin{table*}[!ht]
\centering
\begin{tabular}{l | l | l | l | l | l}
\hline
Model & Hyper-parameter & Value & Model & Hyper-parameter & Value \\ 
\hline\hline
\multirow{8}{*}{BiDAF} & dimension & 300 & \multirow{8}{*}{BERT} & $d_{model}$ & 768 \\
& rnn\_type & LSTM & & num\_heads & 12  \\
& num\_layers & 1 & & num\_layers & 12  \\
& hidden\_size & 300 & & $d_{ff}$ & 3072 \\
& dropout & 0.2 & & dropout & 0.2 \\
& optimizer & Adam & & optimizer & BertAdam \\
& learning rate  & 0.001 & & learning rate  & 0.00003 \\
& batch size  & 16 & & batch size  & 16 \\
& epoch & 15 & & epoch & 5 \\
\hline
\end{tabular}
\caption{Hyper-parameters used in our experiments.
}
\label{table:hyperparameters}
\end{table*}

\begin{table*}[ht]
\centering
\resizebox{\linewidth}{!}{%
\small
\def\arraystretch{1.25}%
\begin{tabular}{p{6.5cm} | p{7cm}}
\hline
Value & Example Question From PolicyQA \\ 
\hline\hline
Collect on website & Do you collect my information on your website?
\\ 
Collect in mobile app & Will you collect my data if I use your phone app?
\\
Collect on mobile website & How do you collect data when I use my mobile?
\\
Track user on other websites & Do they track users' activities on other websites?
\\
Collect from user on other websites & Does the website collect my info on other websites?
\\
Receive from other parts of company/affiliates & Do you collect my information from your affiliates?
\\
Receive from other service/third-party (unnamed)  & Does the website obtain my data from others?
\\
Receive from other service/third-party (named) & Who provides you my data?
\\
Other & How do you receive data from users?
\\
\hline
\end{tabular}
}
\caption{Examples questions from PolicyQA for the ``Action First-Party'' attribute under the \emph{First Party Collection/Use} data practice category.}
\label{table:question_example}
\end{table*}

\end{document}